\documentclass{article}

\usepackage{microtype}
\usepackage{graphicx}
\usepackage{subfigure}
\usepackage{booktabs} 

\usepackage{hyperref}

\usepackage[accepted]{icml2023}

\usepackage{amsmath}
\usepackage{amssymb}
\usepackage{mathtools}
\usepackage{amsthm}

\usepackage[capitalize,noabbrev]{cleveref}

\theoremstyle{plain}

\theoremstyle{definition}

\theoremstyle{remark}

\usepackage[textsize=tiny]{todonotes}

\begin{document}

\twocolumn[
\icmltitle{Brain in the Dark: Design Principles for Neuro-mimetic Learning and Inference}



\icmlsetsymbol{equal}{*}

\begin{icmlauthorlist}
\icmlauthor{Mehran H. Bazargani}{equal,comp}
\icmlauthor{Szymon Urbas}{equal,stats}
\icmlauthor{Karl Friston}{actinf}

\end{icmlauthorlist}

\icmlaffiliation{comp}{School of Computer Science, University College Dublin (UCD), Dublin, Ireland}
\icmlaffiliation{stats}{Department of Mathematics and Statistics, University College Dublin (UCD), Dublin, Ireland}
\icmlaffiliation{actinf}{Wellcome Centre for Human Neuroimaging,  Institute of Neurology, University College London (UCL), London, UK}

\icmlcorrespondingauthor{Mehran H. Bazargani}{mehran.hosseinzadehbazargani@ucd.ie}

\icmlkeywords{Machine Learning, ICML}

\vskip 0.3in
]



\printAffiliationsAndNotice{\icmlEqualContribution} 

\begin{abstract}
Even though the brain operates in pure darkness—within the skull—it can infer the most likely causes of its sensory input. An approach to  modelling this inference is to assume that the  brain \textit{has} a generative model of the world, which it can invert to infer the hidden causes behind its  sensory stimuli, that is, \textit{perception}.  This assumption raises key questions: how to formulate the problem of designing brain-inspired generative models, how to invert  them for the tasks of inference and learning, what is the appropriate loss function to be optimised, and, most importantly, what are the different choices of mean field approximations (MFA) and their implications  for variational inference (VI).
\end{abstract}

\section{Introduction} \label{introduction}
It is remarkable that even though the brain resides in pure darkness in our skull, it is still capable of understanding and analysing the world \textit{out there}, plan for an unseen future and even make decisions that could affect and change the world. For decades, there has been a popular view of the brain as a predictive machine that is constantly inferring the hidden causes behind its sensory inputs.

This view dates back to Helmholtz \cite{helmholtz1866concerning}, who proposed the idea of ``perception as unconscious inference''– a view that has emerged as the “Bayesian brain” hypothesis \cite{doya2007bayesian}. This approach formulates perception as an inferential process based on a generative model of how the brain believes its sensations are generated, where the brain is thought of a statistical organ that updates probabilistic beliefs about the external states of the  world, given the observed sensory data. This formulation appeals to Bayes’ rule, which allows for optimal belief updates, given the sensory stimuli \cite{parr2022active}. More technically, given a sensory observation, $o$, the goal of perception is to infer the most likely hidden cause, $s$, which led to this observation, which can be formulated through the Bayes’ theorem.


In order to define an appropriate generative model—and its inversion, one needs to consider several aspects of the problem at hand. Crucially, one needs to address some foundational questions: are we dealing with continuous or discrete hidden states? And are we looking at continuous or discrete time? Is the task of inference limited to the parameters of the generative model (i.e.~learning), or just the hidden states (i.e.~inference), or both? What is the most suitable objective function—whose optimisation entails learning and inference—and how to extremise it? Does one commit to a functional form for the posteriors? Is there a role for mean-field approximations (MFA)? Should one use sampling schemes or analytic variational inference (VI)? and so on. In this paper, we offer a detailed investigation of these questions and provide a road map towards an accurate and efficient formulation of neuro-mimetic probabilistic generative modelling.

\section{Various problem formulations and their implications}
There are different problem spaces when designing generative models and the method for their inversion. In this section, we will discuss the implications and general considerations to keep in mind before implementing these models.

\subsection{Inference and learning: estimating the hidden states or estimating the parameters}
It is important to clarify if the task of model inversion is in the service of \textit{inference}, i.e.~inferring the most likely distribution over the hidden state (assuming fixed/learned model parameters), given some noisy observations, and/or \textit{learning} the parameters of the generative model as well. Interestingly, the ML community normally focuses on estimating the unknown parameters, where the issue of state estimation is suppressed --- it does not matter if the states are unknown random variables (i.e., with random fluctuations), or whether they are fixed variables conditioned on the parameters (i.e.~a deterministic State Space Model (SSM)).

We use $z$ to denote the collection of all quantities to be inferred (estimated); e.g. $z$ is the set of hidden states and model parameters for inference and learning. The posterior distribution of $z$ based on all observed data, $\mathcal{D}$, is obtained through the Bayes' theorem, which states: $p(z|\mathcal{D})=\frac{p(\mathcal{D}|z)p(z)}{p(\mathcal{D})}$. Apart from special cases, the posterior is not readily available; the normalising constant $p(\mathcal{D})$ can involve a difficult and often high-dimensional sum or integral and has no closed form. This quantity is often referred to as \emph{model evidence}. In the VI framework we wish to identify a surrogate  distribution $q$ which resembles  the true posterior. This approximating posterior is found by using the \emph{variational free energy (VFE)} \cite{friston2010free} defined as 
\begin{align*}
    F(q;\mathcal{D}) &= D_{\mathrm{KL}}[q(z)||p(z)]-\mathbb{E}_{q(z)}[\ln p(\mathcal{D}|z)]\\
                    &= -\mathbb{E}_{q(z)}[\ln p(\mathcal{D},z)/q(z)],
\end{align*}
where $D_{\mathrm{KL}}$ is the Kullback-Leibler divergence.
The VFE quantity is the negative of the evidence lower bound (ELBO).  Variational inference is based on choosing a distribution $q$ from some prespecified class of distributions. Indeed, if, and only if, we had access to the true posterior $p(z|\mathcal{D})$, the VFE would become exactly zero; if we have two distributions, the one closer to the  $p(z|\mathcal{D})$ achieves a lower VFE value --- the minimum of VFE can be a proxy for the intractable model evidence, enabling  Bayesian model selection. This converts the impossible marginalisation problem  into an optimisation problem. As VFE is a functional of $q$ (i.e.~it takes in a function and returns a scalar), calculus of variation is used for minimisation \citep[e.g.~Chapter 10 of][]{BiNa2006}. By inverting the generative model through VFE minimisation one can accomplish: \textbf{(i) unknown parameter estimation}, where there is no interest in hidden-states estimation, and only parameter estimation is of interest (i.e.~learning); and \textbf{(ii) unknown state and parameter estimation}, where model inversion solves a dual estimation problem in partially observed or stochastic systems, where both the hidden states and parameters are estimated (i.e.~inference and learning).

\subsection{State-space model formulations}
For inference, a key question is whether we are working with \textit{discrete states} or \textit{continuous states}. We consider a sequence of states $s_{1:T}$ that we wish to infer, based on potentially noisy observations $o^\tau:=o_{1:T}$. A hidden Markov model (HMM) is characterised by the following properties: (i) $p(s_\tau | s_1, s_2,...,s_{\tau -1})=p(s_\tau | s_{\tau -1})$ and (ii) $p(o_\tau | s_1, s_2,...,s_\tau)=p(o_\tau | s_\tau)$. The former specifies the \emph{Markov transitions} of the hidden states and the latter the \emph{partial observation} process. In Appendix C, Fig.~1 \cite{parr2022active} provides an example of a HMM for inference; this example will be discussed in detail in later sections.

Here, one assumes some particular dynamic generative model composed of: an initial (prior) distribution, \emph{i.e.} $s_1\sim\mu_\theta(\,\cdot\,)$; a transition mechanism, $s_\tau|s_{\tau-1}\sim f_\theta(\,\cdot\,|s_{\tau-1}),~\tau>1$; and an observation (emission) mechanism $ o_\tau|s_\tau \sim g_\theta(\,\cdot\,|s_\tau),~\tau\geq1$; $\theta$ encompasses all model parameters, and we use $\tau$ to simply denote any of the time points where variables are generated ($\tau=1,2,...$). By construction, this generative model is a HMM.

Inference of HMM may concern different posterior distributions: (i) $p(s_{1:\tau}|o_{1:\tau}, \theta)$ (smoothing); (ii) $p(s_\tau|o_{1:\tau}, \theta)$ (filtering); or (iii) $p(s_{\tau+1}|o_{1:\tau},\theta)$ (prediction). Here, we suppose the model parameters could be unknown and thus will need to be included in the inference: to inform parameter learning we will require the whole smoothing distribution (at least in principle). An additional complication arises when we carry out this inference \textit{online}, i.e.~using streaming data: e.g. when the brain continuously assimilates data from the sensorium.

To deploy variational inference, we must decide on a particular form of our MFA, $ q^\psi(s_{1:\tau}, \theta)\approx p(s_{1:\tau}, \theta|o_{1:\tau})$, where the joint distribution $q$ is defined by sufficient statistics or hyperparameters $\psi$; for example, in discrete-state models these would be probability vectors of a categorical distribution, and equivalently the means and variances of a Laplace approximation in continuous-state models \citep[e.g.][]{FrMa2007}.  With the notation $z=(s_{1:\tau},\theta)$, the first factorisation is to separate the latent variables and $\theta$, $q^\psi(s_{1:\tau}, \theta) = q^{\psi}(s_{1:\tau})q^{\psi}(\theta)$. In this general setup, we focus on identifying a $\theta^*$ which maximises the ELBO; this can be read as inference through the \textit{expectation-maximisation} algorithm, or \emph{maximum a posteriori estimation} if a prior for $\theta$ were to be used. In terms of the MFA, we suppose $q^{\psi}(\theta)$ is a Dirac function; we could, however, use a Gaussian family instead \citep{FrMa2007}.

Particular care must be taken when deciding on the MFA form used for the latent variables, $q^{\psi}(s_{1:\tau})$. The ELBO  at time $\tau$, as a function of model and MFA parameters, is 
\begin{equation*}
    L_\tau(\theta,\psi):=\mathbb{E}_{q^\psi(s_{1:\tau}|o^\tau)}\left[\ln\left(p(s_{1:\tau}, o_{1:\tau}|\theta)/q^\psi(s_{1:\tau}|o^\tau)\right)\right];
\end{equation*}
this is our objective function based on all the information available at that time under a particular form for the MFA. The goal of the scheme is to obtain $\arg\max_{\theta,\psi} L_\tau(\theta,\psi)$ at each time point $\tau$ as the data appear. If $\psi$ and $\theta$ are specified to be real-valued, this will require the gradients $\nabla_\theta L_\tau(\theta,\psi)$ and $\nabla_\psi L_\tau(\theta,\psi)$. 

The ensuing Bayesian belief updating for filtering or smoothing will have different functional forms depending on the type of state space in question: continuous states or discrete states.

\subsubsection{Discrete State Space Models}
Suppose the hidden state, $s_\tau$, can take $K$ possible values. In this case, the initial prior probability regarding the hidden state is encoded in a $K-$dimensional vector $D$ expressed as a categorical distribution at time $\tau = 1$ as $P(s_1)$. Since both the states and outcomes are categorical variables, the likelihood has also a categorical distribution, parameterised by the $K\times K$ matrix \textbf{A}: $P(o_\tau) = \mathsf{Cat}(\textbf{A})$ where $A_{ij} = P(o_\tau = i|s_\tau = j)$. Furthermore, the transition probability for the states is parameterised by the $K \times K$ matrix \textbf{B}: $P(s_{\tau + 1} | s_\tau) = \mathsf{Cat}(\textbf{B}_\tau)$. Finally, one can compute the prior over the sequence of hidden states, denoted as $\tilde{s}$, using the prior over the initial state, expressed by vector $D$, and state transition beliefs denoted by the matrix $\textbf{B}$: $P(\tilde{s})=P(s_1)\prod_{\tau=1}^{} P(s_{\tau+1}| s_\tau)$. The prior, likelihood, and state transitions probabilities together, constitute the HMM generative model for leaning and inference. The learning is possible since by defining priors over the parameters of the model, we can now update these belief (See Fig.~2 in Appendix C). Here, we present a finite state-space but the methods covered in this work could be applied to infinite state-space problems with an appropriate transition distributions motivated by an assumed data-generating model (e.g.~random walk on all integers, $s_\tau\in\mathbb{Z}$); the VFE construction would involve the same steps.

\subsubsection{Continuous State Space Models}
As opposed to the discrete SSM, in a continuous state space model, both the states and observations can take real continuous values. Using the following pair of stochastic equations, at a given time $\tau$, we can determine how hidden states, $s_\tau$, generate observation, $o_\tau$, and how states evolve over time, as parametrised by $\nu_\tau$:
\begin{align*}
\dot{s}_\tau = f(s_\tau, \nu_\tau) + \omega_\tau^1~~~\mbox{and}~~~    o_\tau = g(s_\tau) + \omega_\tau^2,
\end{align*}
where  $\dot{s}_\tau$ is the first-order time derivative of the hidden state $s_\tau$, representing the rate of change of the hidden state. Furthermore, $\omega_\tau^1$ and $\omega_\tau^2$ represent the random fluctuations corresponding to the states and observations, respectively; the two random processes are assumed to be independent (e.g.~\citep{friston2010generalised}). In the most basic case, these could be Wiener processes, with independent increments, but other smoother processes such as the Mat\'ern process could be used here \citep[e.g.][]{HaSa2010} The first equation describes the evolution of hidden states over time through a deterministic function $f(s_\tau, \nu_\tau)$ and stochastic fluctuations $\omega_\tau^1$. Here, we suppose the evolution of the hidden states can be modelled as differential equations, i.e.~the change from $\tau=1$ to $\tau=2$ comprises infinitesimally small increments in time. The second equation expresses how the observations are believed to be generated from the hidden state. Interestingly, if we assume the fluctuations to be normally distributed, these two equations form a generative model that underwrites Kalman-Bucy filter \citep[e.g.][]{ruymgaart2013mathematics} in engineering literature.

It is interesting to note that, even though we are observing this continuous state space model at discrete times, the underlying dynamics of the system are continuous in time (e.g.~the evolution of the hidden states, VFE minimisation, etc.). By collapsing the hidden states and their motion into one state variable $\Breve{s}_\tau=\{\dot{s}_\tau, s_\tau\}$, the approximate posterior, $q$, can now be written as $q(\Breve{s}_\tau)$ where $\Breve{s}_\tau$ is now an augmented variable. Then the standard VFE can be derived and minimised in the usual way, during the time intervals between observations. More specifically, after observing $o_\tau$, we can minimise the integration of point estimates of VFE along a continuous time interval, $T$, until the next observation $o_{\tau + T}$. This quantity is called \textit{Free Action} and is defined as $\overline{\mathcal{A}}[q(\Breve{s})]=\int_{\tau}^{\tau+T} \mathrm{VFE}[q(\Breve{s}_t)] \,\mathrm{d}t$, and it is an upper bound on the accumulated surprise, $-\ln(\overline{P}(o))=-\int_{\tau}^{\tau+T} \ln(P(o_t)) \,\mathrm{d}t$, over the same time period. Thus, by minimising $\overline{\mathcal{A}}$ in-between observations, the generative model is constantly minimising VFE of a path of length $T$, and thus continuously striving to improve the estimation of the posterior over the hidden states and/or parameters.

Interestingly, random fluctuations in the data-generating mechanism, $\omega$, are generally assumed to have uncorrelated increments over time (i.e.~Wiener assumption), however, in most complex systems (e.g.~biological systems)---where the random fluctuations themselves are generated by some underlying dynamical ---they possess a certain degree of smoothness. Indeed, by relaxing the Wiener assumption and imposing smoothness on the model functions $f$ and $g$, we have the opportunity to not only consider the rate of change of the hidden state and the observation, but also their corresponding higher order temporal derivatives (i.e.~acceleration, jerk, etc.); see, for example, \cite{friston2010generalised}. The resultant pair of $\{s, \dot{s}, \ddot{s},...\}$ and $\{o, \dot{o}, \ddot{o},...\}$ are called the \textit{generalised coordinates of motion} \cite{balaji2011bayesian}, which provides an opportunity for further capturing the dynamics that govern the evolution the hidden states and observations. An estimated trajectory over time can be calculated using a Taylor series expansion around the present time, which results in a function that can extrapolate to the near future as well as the recent past. Now, the mapping from continuous to discrete time is possible using this expansion, where one can map from the generalized coordinates of motion to the discretised time.

\subsection{Perception modelling: online variational inference in practice}

Let us now examine some choices for the MFA and practical considerations that may arise. The simplest, and perhaps naive, choice is the fully decoupled factorisation $q^{\psi}(s_{1:\tau}) = \prod_{t=1}^\tau q^{\psi}_t(s_t)$. However, a more natural option --- motivated by the Markov process ---- is $q^{\psi}(s_{1:\tau}) = q^{\psi}_1(s_1)\prod_{t=2}^\tau q^{\psi}_t(s_t|s_{t-1})$. Unfortunately, these two MFAs do not readily lend themselves to online inference: the ELBO in these cases involve an integral over the true filtering distributions $p_\theta(s_{t-1}|o_{t-1})$ which itself has no closed form \citep[Section 2 of][]{ZhPa2020}. Instead, an approximation to the ELBO can be used: $\widehat{L}_\tau(\theta,\psi) = \sum_{t=1}^\tau\mathbb{E}_{q_t^\psi(s_{t-1},s_t)}\left[\ln \frac{f_\theta(s_t|s_{t-1})g_\theta(o_t|s_t)}{q_t^\psi(s_{t-1},s_t)}\right]$; this allows for gradient calculations at a constant computational cost at each time point. Where $L_\tau$ is an expectation over the full approximate smoothing distribution of $s_{1:\tau}$, $\widehat{L}_\tau$ is instead a sum of pairwise expectations. However, it can be shown that $\widehat{L}_\tau\leq L_\tau$ --- VFE will not be truly minimised. One way around that is to employ a \emph{reversed} version of the MFA, $q^{\psi}(s_{1:\tau}) = q^{\psi}_\tau(s_\tau)\prod_{t=1}^\tau q^{\psi}_t(s_{t-1}|s_t)$; the formulation is for the mathematical convenience, and does not change the original HMM. As outlined in Proposition 1 of \cite{CaSh2021}, the ELBO under this MFA has a recursive form: $L_\tau(\theta,\psi) = \mathbb{E}_{q_\tau^\psi(s_\tau)}[V_\tau^{\theta,\psi}(s_\tau)]$, where $V_\tau^{\theta,\psi}(s_\tau) = \mathbb{E}_{q^\psi(s_{1:\tau-1}|s_\tau)}[\ln p(s_{1:\tau},o^\tau|\theta)/q^\psi(s_{1:\tau})]$; it is free of the problematic integral which appears for other MFA options, and $V_\tau$ can be expressed in terms of $V_{\tau-1}$ --- VFE calculations at time $\tau$ reuse the quantities from the previous time point. The MFA parameters are indexed with a subscript; calculations may involve the current iterations MFA, $\psi_\tau$, as well as the previous iteration $\psi_{\tau-1}$.

We now revisit the discrete state-space example. In this setting, each component of the MFA, $q_t(s_t)$, $t=1,\ldots, \tau$, at time $\tau$, is a categorical distribution with a probability vector $\boldsymbol{\pi}_\tau^t = (\pi_\tau^t(1),...,\pi_\tau^t(K))$\footnote{The expectations with respect to distributions $q_t$ will take the form of summations; \emph{e.g.} $\mathbb{E}_{q^\psi(s_t| o^\tau)}[h(s_t)]=\sum_{k=1}^K \pi_\tau^t(k) h(k)$, for some integrable function $h$.}, where $P(s_t=k|q_t,o_\tau)=\pi_\tau^t(k)$, the probability of the hidden state at time $t\leq\tau$ having value $k$, conditional on the information available at time $\tau$; Appendix \ref{app:parameter} gives the exact parametrisation for this vector through $\psi$. The transition matrix $\mathbf{B}$ and emission matrix $\mathbf{A}$ follow similar notation where $[\mathbf{B}]_{ij} = \beta^i(j)$ and $[\mathbf{A}]_{ij} = \alpha^i(j)$; Appendix \ref{app:parameter} details the parameterisation of the model through $\theta$. 

Learning and inference based on streaming data, through optimising the exact ELBO (or equivalently VFE) involves gradient-based updates in between the receipt of packets of data; e.g.~suppose, $5$ seconds after receiving $o_\tau$, $o_{\tau+1}$ appears, and that this permits $80$ updates on $\psi$ followed by $50$ updates on $\theta$. Using the reverse version of MFA, the gradients in the updates on $\psi$ will have the form $\nabla_\theta L_\tau(\theta,\psi) = \sum_{l=1}^K \pi_\tau^\tau(l)U_\tau^{\theta,\psi}(l)$, which is calculated by recursion 
\begin{align*}
    U_{t}^{\theta,\psi}(l) &= \sum_{k=1}^K \pi_{t}^{t-1}(k)\left[U_{t-1}^{\theta,\psi}(l) + u_{t}^\theta(k,l)\right],\\
    U_1^{\theta,\psi}(l) &= \nabla_\theta \ln \mu(l)g_\theta(o_1|l)= \nabla_\theta \ln{\alpha}^l(o_1), \\
    u_{t}^\theta(k,l) &= \nabla_\theta\ln f_\theta(l|k)g_\theta(o_{t}|l)= \nabla_\theta\ln {\beta}^k(l){\alpha}^l(o_{t}),
\end{align*}
where $t=2,\ldots,\tau$ \citep[Proposition 2 of][]{CaSh2021}.

To focus on the online inference, we only update $q_\tau$ and $q_{\tau-1}$ when a new observation comes in at time $\tau$.\footnote{For example, at time $\tau=4$, we infer the current hidden state, $s_4$ and use this information to improve our hidden-state inference for the previous time point, $s_3$. Then at time $5$, we infer $s_5$ and improve inference on $s_4$, without changing the posterior approximation of $s_3$; and so on.} This is akin to only updating our short-term memory along with what we currently perceive, leaving the long-term memory fixed. The gradient of the ELBO with respect to the state-space MFA parameters can also be computed recursively, $\nabla_{\psi_\tau} L_\tau(\theta,\psi) = \nabla_{\psi_\tau}\sum_{l=1}^K \pi_\tau^\tau(l)V_\tau^{\theta,\psi_{1:\tau}}(l)$, where
\begin{align*}
    &V_{t}^{\theta,\psi_{1:t}}(l) = \sum_{k=1}^K \pi_{t}^{t-1}(k)\left[V_{t-1}^{\theta,\psi_{1:(t-1)}}(k) + v_{t}^{\theta,\psi_{1:t}}(k,l)\right],\\
    &V_{1}^{\theta,\psi_{1}}(l) = \ln \mu(l)g_\theta(o_1|l)
    = \ln \mu(l){\alpha}^l(o_1),\\
    &v_{t}^{\theta,\psi_{1:t}}(k,l)=\ln \frac{f_\theta(l|k)g_\theta(o_{t}|l)}{m_{t}^{\psi_{1:t}}(l|k)}=\ln \frac{{\beta}^k(l){\alpha}^l(o_{t})}{m_{t}^{\psi_{1:t}}(l|k)},
\end{align*}
where $t=2,\ldots,\tau$, and the conditional $m_{t}^{\psi_{1:t}}$ quantity is detailed in Appendix \ref{app:mfa}.


\section{Conclusion}
A generative model can help us model our beliefs about the data generating process in the world, given uncertain observations. It is by inverting the generative model that we can estimate: 1) the hidden states that cause these observations, and 2) the parameters of the generative model to explain how observations are caused.
This paper offers a comprehensive guide on designing and inverting generative models for both inference and learning, as well as loss function selection and most importantly, different choices of mean-field approximation (MFA) for variational inference. We have illustrated the discrete SSM; however, the foundational concepts are transferable to the continuous SSM.

\section{Future Work}
We are planning to develop a brain-inspired (neuro-mimetic) framework for inverting generative or world models, for the task of perception. The framework of choice is called Predictive Coding (PC), which provides a powerful mathematical framework for describing how the cortex extracts information from noisy stimuli \cite{huang2011predictive}. PC assumes that the brain entails a generative model of the world, under which it constantly makes predictions about the hidden causes behind sensory inputs.
PC is a special case of variational inference where it is assumed that the mean-field factors and posterior probabilities follow Gaussian and Dirac distributions, respectively. Because PC can be formulated as variational inference --- and VFE provides a bound on model evidence---one can use Bayesian model comparison to evaluate different MFA factorisations.

We are also aiming to finesse the variational inference process by capturing higher-order temporal derivatives of hidden states and observations in generalised coordinates of motion. Using this generalised dynamics, variational inference can, in principle, provide a more accurate and efficient estimation of the true posterior over the hidden states, especially in on-line learning under analytic (i.e.~smooth) random fluctuations.

\section*{Acknowledgements}
Mehran H. Bazargani is supported by Enterprise Ireland and the Department of Business, Enterprise and Innovation through the Disruptive Technologies Innovation Fund (DT 2018 0185A) and the Science Foundation Ireland through the Insight Centre for Data Analytics (12/RC/2289\_P2).

Szymon Urbas is supported by funding from Science Foundation Ireland and the Department of Agriculture, Food and Marine on behalf of the Government of Ireland under Grant Number [16/RC/3835] - VistaMilk.

Karl Friston is supported by funding for the Wellcome Centre for Human Neuroimaging (Ref: 205103/Z/16/Z), a Canada-UK Artificial Intelligence Initiative (Ref: ES/T01279X/1) and the European Union’s Horizon 2020 Framework Programme for Research and Innovation under the Specific Grant Agreement No. 945539 (Human Brain Project SGA3).


\bibliography{example_paper}
\bibliographystyle{icml2023}

\newpage
\appendix
\onecolumn

\section{MFA changing with time}\label{app:mfa}
In the setting of streaming data the mean-field approximation is being augmented each time new data arrives. We consider an MFA of the form 
\[
    q^\psi(s_{1:\tau}) = q_\tau^\psi(s_{\tau})\prod_{t=1}^{\tau-1} q_t^\psi(s_{t}|s_{t+1})
\]
which allows an update from $q^\psi(s_{1:\tau})$ to $q^\psi(s_{1:\tau+1})$ via
\[
    q^\psi(s_{1:\tau+1}) = q^\psi(s_{1:\tau}) m^\psi_{\tau+1}(s_{\tau+1}|s_\tau)~~~\mbox{where}~~~m_{\tau+1}^\psi(s_{\tau+1}|s_\tau) = \frac{q_{\tau+1}^{\psi}(s_\tau|s_{\tau+1})q_{\tau+1}^{\psi}(s_{\tau+1})}{q_\tau^{\psi}(s_\tau)}.
\]
In the online inference of the discrete state-space model considered in the main article, we update the MFA hyperparameters, that is, there is no single $\psi$ used throughout and instead we have a sequence $\psi_{1:\tau}$ which itself is augmented at each time point. The gradients of the ELBO (and equivalently VFE) will involve the conditional quantities
\begin{align*}
    m_{t+1}^{\psi_{1:t+1}}(s_{t+1}|s_t) &=\frac{q_{t+1}^{\psi_{1:t+1}}(s_t|s_{t+1})q_{t+1}^{\psi_{1:t+1}}(s_{t+1})}{q_t^{\psi_{1:t}}(s_t)}\qquad (t=1,...,\tau-1)\\&=\frac{\pi_{t+1}^t(s_t)\pi_{t+1}^{t+1}(s_{t+1})}{\pi_{t}^t(s_t)},
\end{align*}
where the second equality follows for the discrete state space model.

\section{Discrete state space model parametrisation}\label{app:parameter}
In order to allow gradient-based updates on the parameters governing the mean-field approximation $q^\psi(s_{1:\tau})$, we use the following parameterisation: $\pi_\tau^t(k)=\exp(\rho_\tau^t(k))/\sum_l\exp(\rho_\tau^t(l))$, $t=1,...,\tau$, where $\rho_\tau^t(1)=0$  and $\rho_\tau^t(k)\in\mathbb{R}$ for all $\tau,~t$ and $k\neq1$. The constraint on the first element of the row vector ensures identifiability. With this notation, the MFA parameters at time $\tau$ are $\psi_\tau=(\boldsymbol{\rho}_\tau^1,...,\boldsymbol{\rho}_\tau^\tau)^\top$.

The transition matrix $\mathbf{B}$ and emission matrix $\mathbf{A}$ follow similar a parametrisation where $[\mathbf{B}]_{ij} = \beta^i(j) = \exp(\tilde{\beta}^i(j))/\sum_k\exp(\tilde{\beta}^i(k))$ and $[\mathbf{A}]_{ij} = {\alpha}^i(j) = \exp(\tilde{\alpha}^i(j))/\sum_k\exp(\tilde{\alpha}^i(k))$, with $\tilde{\beta}^i(1)=0$ and $\tilde{\alpha}^i(1)=0$, and $\tilde{\alpha}^i(k),\tilde{\beta}^i(k)\in\mathbb{R}$ for $k=2,...,K$, for all $i$. We suppose a fixed initial prior on the process $s_t$, that is, $s_1\sim\mu$ which itself is free of $\theta$. The model parameters are $\theta = \left(\tilde{\boldsymbol\alpha}^1,...,\tilde{\boldsymbol\alpha}^K,\tilde{\boldsymbol\beta}^1,...,\tilde{\boldsymbol\beta}^K\right)^\top$.

\section{HMM for inference/learning}
The HMM in Fig.1, represents the evolution of a sequence of hidden states, $s_\tau$, over time. At each time step, $\tau$, a hidden state emits an observation, $o_\tau$, and the state at any one time depends \textit{only} on the state at the previous time where this dependency is encoded in the matrix \textbf{B}. The initial prior probability regarding the hidden state is encoded in the vector $D$, and finally, the matrix \textbf{A} encodes the likelihood distribution of generating outcomes under each state \cite{parr2022active}. Here, it is assumed that the parameters of the generative model is learned and we are only interested in inferring the hidden states.

\begin{figure}[h]\label{discrete_hmm}
\centering
\includegraphics[width=.45 \textwidth]{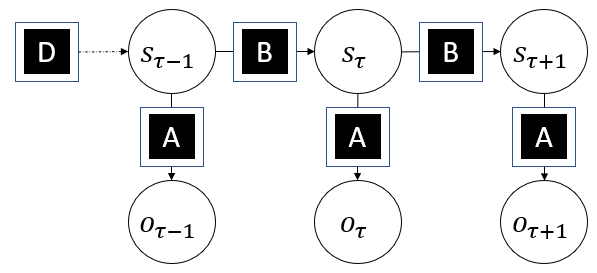}
\caption{A Hidden Markov Model (HMM) for inference.}
\end{figure}
\newpage
The HMM in Fig.2, represents the evolution of a sequence of hidden states, $s_\tau$, over time, with priors over the parameters of the model, \textbf{A}, \textbf{B}, and $D$ \cite{parr2022active}. Here, we are interested in inferring the hidden states and learning the parameters of the generative model.

\begin{figure}[ht]\label{discrete_hmm_learning parameters}
\centering
\includegraphics[width=.65 \textwidth]{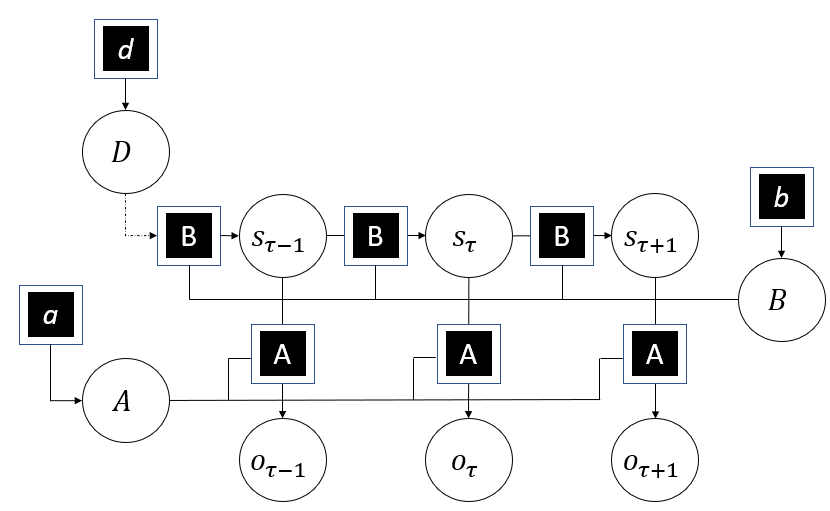}
\caption{A Hidden Markov Model (HMM) for learning and inference.}
\end{figure}

\end{document}